\documentclass[letterpaper]{article} 
\usepackage{aaai2026}  
\usepackage{pifont}
\usepackage{comment}
\usepackage{times}  
\usepackage{helvet}  
\usepackage{courier}  
\usepackage[hyphens]{url}  
\usepackage{graphicx} 
\usepackage{xspace}
\usepackage{tabularx}

\usepackage{listings}
\usepackage{xcolor}  

\usepackage{natbib}  
\usepackage{caption} 
\usepackage{todonotes}
\usepackage{algorithm}

\usepackage{amsmath}
\usepackage{algorithmic}
\usepackage{xcolor}
\definecolor{darkgreenaccent}{RGB}{112, 173, 71}  
\definecolor{darktealaccent}{RGB}{68, 114, 196}   

\definecolor{precblue}{RGB}{0,0,255} 
\definecolor{recgreen}{RGB}{0,128,0} 
\definecolor{f1red}{RGB}{255,0,0} 

\definecolor{yesgreen}{RGB}{0,128,0} 
\definecolor{nored}{RGB}{255,0,0} 
\usepackage{romannum}
\newcommand{\ie}{i.e.,\xspace}
\usepackage{newfloat}
\usepackage{booktabs}
\usepackage{array} 
\usepackage{listings}
\DeclareCaptionStyle{ruled}{labelfont=normalfont,labelsep=colon,strut=off} 
\floatstyle{ruled}
\newfloat{listing}{tb}{lst}{}
\floatname{listing}{Listing}
\title{Audited Reasoning Refinement: Fine-Tuning Language Models\\ via LLM-Guided Step-Wise Evaluation and Correction}
\author{
    Sumanta Bhattacharyya,
    Sara Riazi,
    Pedram Rooshenas
}
\affiliations{
    Department of Computer Science\\
    University of Illinois Chicago\\
    \{sbhatt54,riazi,pedram\}@uic.edu
}

\usepackage{listings}
\usepackage{xcolor}

\lstdefinelanguage{json}{
    basicstyle=\ttfamily\small,
    numbers=left,
    numberstyle=\tiny\color{gray},
    stepnumber=1,
    numbersep=8pt,
    showstringspaces=false,
    breaklines=true,
    frame=single,
    literate=
     *{0}{{{\color{numb}0}}}{1}
      {1}{{{\color{numb}1}}}{1}
      {2}{{{\color{numb}2}}}{1}
      {3}{{{\color{numb}3}}}{1}
      {4}{{{\color{numb}4}}}{1}
      {5}{{{\color{numb}5}}}{1}
      {6}{{{\color{numb}6}}}{1}
      {7}{{{\color{numb}7}}}{1}
      {8}{{{\color{numb}8}}}{1}
      {9}{{{\color{numb}9}}}{1}
      {:}{{{\color{punct}{:}}}}{1}
      {,}{{{\color{punct}{,}}}}{1}
      {\{}{{{\color{delim}{\{}}}}{1}
      {\}}{{{\color{delim}{\}}}}}{1}
      {[}{{{\color{delim}{[}}}}{1}
      {]}{{{\color{delim}{]}}}}{1},
}

\lstdefinelanguage{prompt}{
    basicstyle=\ttfamily\small,
    numbers=left,
    numberstyle=\tiny\color{gray},
    stepnumber=1,
    numbersep=8pt,
    showstringspaces=false,
    breaklines=true,
    frame=single,
    literate=
     *{0}{{{\color{numb}0}}}{1}
      {1}{{{\color{numb}1}}}{1}
      {2}{{{\color{numb}2}}}{1}
      {3}{{{\color{numb}3}}}{1}
      {4}{{{\color{numb}4}}}{1}
      {5}{{{\color{numb}5}}}{1}
      {6}{{{\color{numb}6}}}{1}
      {7}{{{\color{numb}7}}}{1}
      {8}{{{\color{numb}8}}}{1}
      {9}{{{\color{numb}9}}}{1}
      {:}{{{\color{punct}{:}}}}{1}
      {,}{{{\color{punct}{,}}}}{1}
      {\{}{{{\color{delim}{\{}}}}{1}
      {\}}{{{\color{delim}{\}}}}}{1}
      {[}{{{\color{delim}{[}}}}{1}
      {]}{{{\color{delim}{]}}}}{1},
}

\definecolor{numb}{rgb}{0.37,0.37,0.37}
\definecolor{punct}{rgb}{0.6,0.6,0.6}
\definecolor{delim}{rgb}{0.2,0.2,0.2}
\newcommand{\incorrect}[1]{\textcolor{red}{#1}}

\newcommand{\correct}[1]{\textcolor{green}{#1}}

\usepackage{bibentry}

\begin{document}

\maketitle

\begin{abstract}
Training a task-specific small reasoning model is challenging when direct human supervision or high-quality labels are scarce. However, LLMs with reasoning capabilities produce abundant intermediate reasoning traces that can be systematically refined to create effective supervision signals. We propose Reason-Refine-then-Align (R$^2$tA), which turns refined model rationales into supervision for training task-specific reasoning models. Our method generates initial reasoning and responses from an open-source base model on task-specific inputs, then refines these traces, fixing hallucinations and inconsistencies, to form a high-fidelity dataset. We perform a two-stage alignment, supervised fine-tuning (SFT), followed by direct preference optimization (DPO) to calibrate the model's intermediate reasoning with human-validated conceptual preferences and then condition the final output on that aligned reasoning. As a case study, we apply R$^2$tA to evaluate extended entity relationship diagrams (EERDs) in database system design, a structurally complex task where prompt-only methods miss or hallucinate errors. We curated a dataset of 600 EERD variants (train/test split of 450/150, respectively) with induced mistakes spanning 11 categories. Empirical evaluation suggests R$^2$tA provides a practical, cost-effective path to scalable LLM adaptation in data-scarce domains, enabling reproducible AI tools for education and beyond.

(R$^2$tA)

\end{abstract}


\section{Introduction}

Training task-specific reasoning models, particularly in data-scarce and high-stakes domains like education~\cite{yan2024practical} and healthcare~\cite{singhal2023large}, faces significant hurdles in achieving reliability and transparency. Existing LLM adaptation strategies, such as prompt engineering and self-generated synthetic labels, are brittle, prone to bias, and frequently miss or hallucinate errors in complex reasoning tasks~\cite{wang-etal-2023-self-instruct,li-etal-2023-synthetic,zhu2023promptrobust,guo2024makes}. This brittleness suggests traditional accuracy metrics often overestimate the performance of NLP models, uncovering critical failures even in state-of-the-art systems~\cite{ribeiro2020beyond}. Furthermore, models lack robustness to real-world noise, decreasing performance even when small changes are introduced~\cite{moradi2021evaluating}. Meanwhile, stakes are exceptionally high in education, where a faulty hint can misguide thousands of learners. Recent work confirms that LLM-generated explanations for programming errors are not yet reliable~\cite {phung2023generating}. \par

While LLMs can generate initial reasoning traces, prior work shows they often perform poorly in directly evaluating complex structures~\cite{turpin2023language}, leading to inconsistencies, hallucinations, and missed domain nuances in data-scarce settings~\cite{chalkidis2022lexglue}. However, LLMs are frequently better at verifying or checking an answer than producing the right one on the first try~\cite{zhang2024small,li2022making}, enabling them to serve as verifiers through techniques like self-consistency~\cite{wang2022self}, backward verification~\cite{weng2023large,jiang2023forward}, and critique-refine loops~\cite{zhang2024small}. This motivates a general, label-light recipe that leverages LLMs as verifiers to refine the reasoning and iteratively trace. Reward signals encoding high-level conceptual preferences steer each refinement step, turning verification into trustworthy supervision for task-specific reasoning models.\par

\begin{figure*}[!t]
  \centering
  \includegraphics[
    width=\textwidth,
  ]{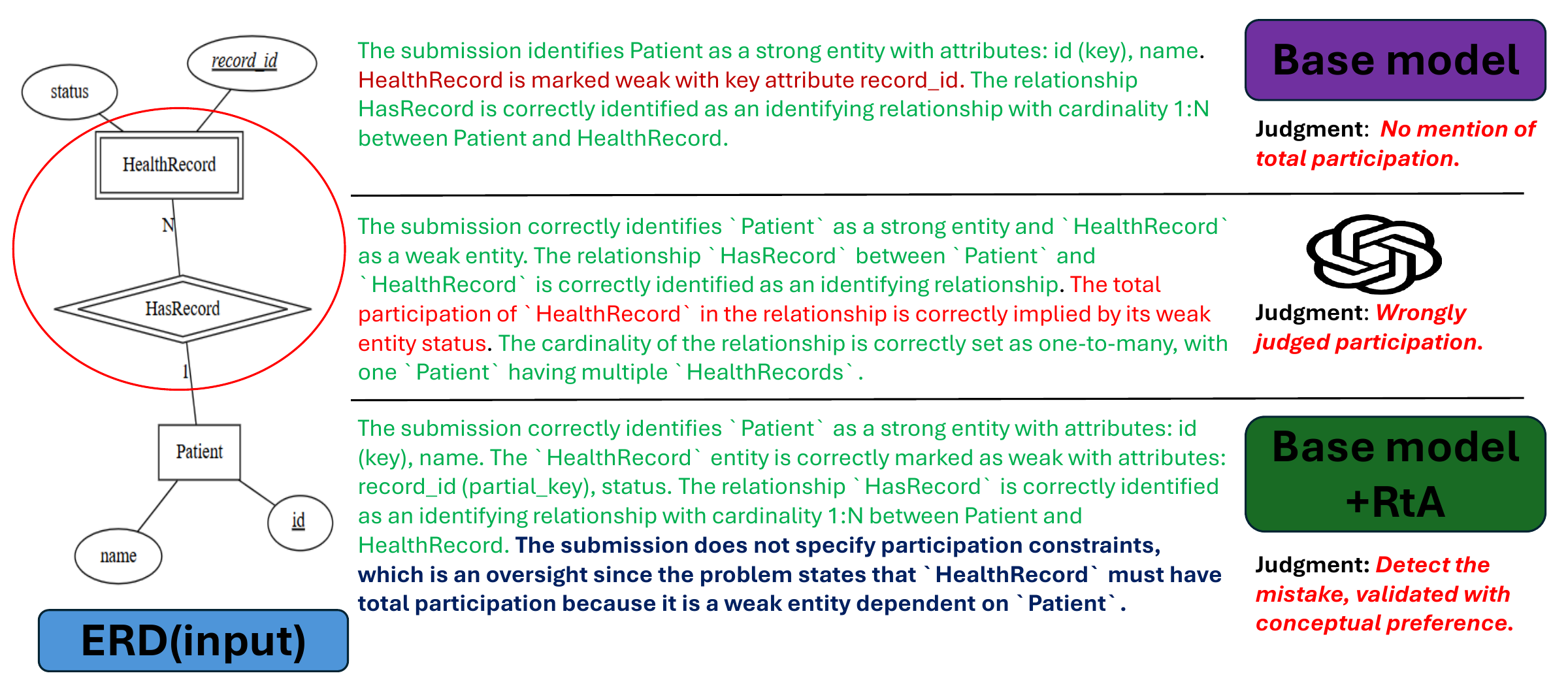}
  \caption{Comparative evaluation of methods on EERD error detection. The input is provided as JSON format, contains an intentionally introduced error in total participation (highlighted in red circle). The error violates a fundamental database principle: weak entities must participate totally in their identifying relationships with strong entities. Specifically, the HealthRecord weak entity should have total participation (should be double line) in the HasRecord identifying relationship with its corresponding strong entity. The judgment section presents each method's response to this conceptual error. Results demonstrate that R$^2$tA successfully identifies the mistake through enhanced reasoning validated against conceptual preferences, while other methods either fail to detect the error(base model) or provide incorrect assessments (GPT-4o). }
  \label{fig:highlevel}
\end{figure*}


 Moreover, supervised fine-tuning (SFT)~\cite{radford2018improving} can degrade Chain-of-Thought (CoT) ~\cite{wei2022chain} and compromise the faithfulness of generated rationales, yielding unreliable use of intermediate steps for final outputs \cite{lobo2025impact}. Prior work~\cite{paul2024making} studies causal links between CoT and answers in generic QA, but not requirement‑grounded, mistake‑level diagnostics that must yield actionable feedback under domain constraints. Recent work shows that verifying each intermediate step \cite{li2023making}, 
reconciling multiple reasoning paths through self-consistency \cite{wang2022self}, and grounding events 
in a temporal narrative \cite{zhang2024narrative} markedly reduce reasoning errors, thereby motivating structured 
refinements that ensure mistake-level coverage and actionable feedback.
 

 We introduce \textbf{Reason-Refine-then-Align (R$^2$tA)}, a framework that systematically refines LLM-generated intermediate reasoning traces. It leverages established superior verification capabilities of LLMs over the initial generation to create high-fidelity supervision signals without relying on extensive human labeling. R$^2$tA first obtains reasoning and responses from an open-source base model on task-specific inputs. It then employs a guide LLM to iteratively refine these traces in two stages: (\romannum{1}) technical refinement based on precision and recall rewards, iterating until maximum values are reached to ensure the reasoning is technically sound; (\romannum{2}) writing refinement using rewards for coherence and chronological relevance, preserving precision/recall while structuring the text for better logical flow and readability. These refinements verify conceptual preferences via rewards, fixing hallucinations, inconsistencies, and missed errors to produce polished, domain-aligned rationales. R$^2$tA then fine-tunes the model via a two-stage alignment: SFT to calibrate intermediate reasoning, followed by DPO~\cite{rafailov2023direct} to align it with conceptual preferences and condition final outputs on that aligned reasoning. \par


  
  As a case study, we demonstrate \textbf{R$^2$tA} on the Extended Entity–Relationship Diagram (EERD) evaluation in database system education~\cite{riazi2025llm}. EERD evaluation stands out for task-specific reasoning models due to its graph-like dependencies (\ie entities, relationships etc) requiring consistent, multi-step traces where errors cascade, ideal for testing R$^2$tA's iterative refinements. It demands granular diagnostics across multiple mistake types, exposing LLM verification brittleness in data-scarce, high-stakes education, enabling R$^2$tA to yield transparent, faithful models for structured domains beyond linear tasks. Our key contributions include:
\begin{itemize}
    \item \textbf{ Reason-Refine-then-Align (R$^2$tA):} a general, label-light framework that converts refined LLM rationales into supervision for training task-specific reasoning models, decoupling reasoning refinement from output alignment.
    
    \item We curate a \textbf{high-fidelity dataset of 600 EERD variants with induced errors spanning 11 categories, along with a detailed curation protocol}, providing a reproducible resource for structured reasoning tasks in education and beyond.
    \item We empirically validate \textbf{R$^2$tA} on the challenging task of EERD evaluation in database education, demonstrating superior performance.
    
\end{itemize}

\section{Method}

\begin{figure*}[!t]
  \centering
  \includegraphics[
    width=0.7\textwidth,
    keepaspectratio
  ]{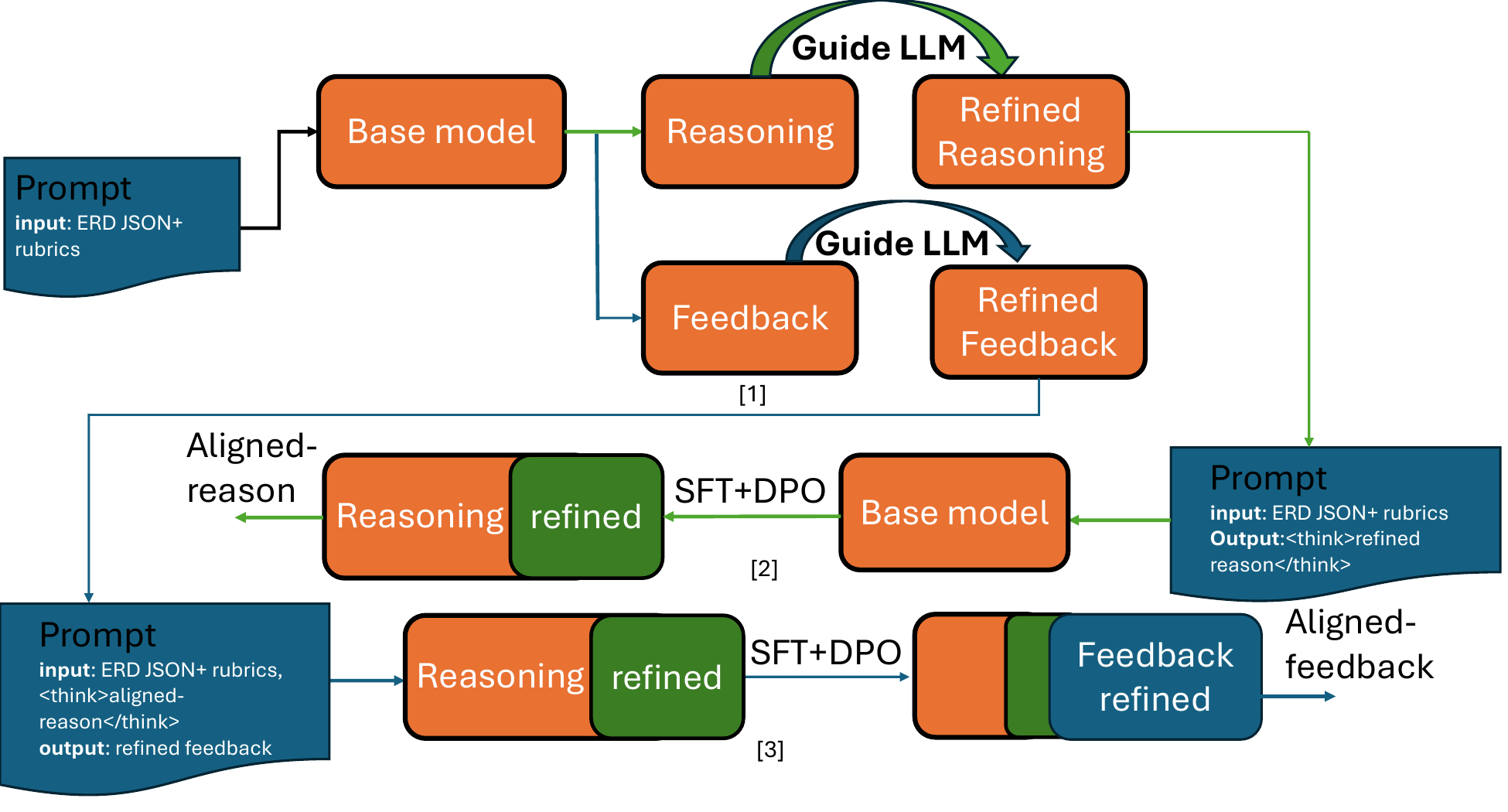}
  \caption{This figure summarizes our entire pipeline (Algorithm \ref{alg:rta}). The Guide LLM refines reasoning and feedback against validated conceptual preferences detailed in Algorithm \ref{alg:factaudit} and Algorithm \ref{alg:stylepolish}, respectively.} 
  \label{fig:fulllevel}
\end{figure*}

A single training step of our proposed R$^2$tA algorithm is broken down as follows. Algorithm \ref{alg:rta} and Figure \ref{fig:fulllevel} still do the explanation.

\subsection{Background}
Data-scarce domains need faithful CoT reasoning in LLMs, with outputs conditioned on domain-aligned rationales.
Fine-tuning often erodes this faithfulness, creating inconsistent traces. R$^2$tA's recipe uses guided LLM auditing to verify and correct traces for high-fidelity supervision. SFT calibrates reasoning coherence, followed by DPO aligns preferences and conditions outputs. This audited, decoupled process unlocks label-efficient adaptation.

\subsection{Refinement}

Minor inconsistencies in EERD evaluation can cascade into misleading educational feedback due to the domain's constraint-heavy nature. We developed a two-phase refinement pipeline that splits factual checks from style tweaks. It ensures factual accuracy, comprehensive coverage of mistake types, and enhanced readability, decoupling technical auditing from stylistic polishing.

\subsubsection{Reasoning Trace Auditing}

The primary objective of this stage is to convert noisy, potentially hallucinatory CoT traces into audited versions that are (\romannum{1}) factually aligned with task constraints and (\romannum{2}) exhaustive at the mistake-type granularity essential for actionable diagnostics.

For each input $x \in \mathcal{D}_x$, we first prompt the base LLM $\pi_0$ to generate an initial reasoning trace $r_0$ and output $y_0$. Guide LLM $G$ audits iteratively (See Algorithm~\ref{alg:factaudit} ), evaluating $r_0$ against domain-specific rubrics to identify false positives (hallucinated claims) and false negatives (omitted errors). Hallucinations are surgically excised by removing erroneous phrases, while missed mistakes are rectified through the insertion of concise, tailored explanations. After each step of refinement, we compute F1 score to quantify coverage of ground-truth errors. The process converges when the F1 score stabilizes (\ie the absolute difference between consecutive scores falls below a threshold $\epsilon$, with saturation checked over two iterations to avoid premature halting). This yields a refined trace $r$ with an optimal F1 score $F1_{optimal}$, ensuring factual alignment without over-refinement (see Stage 1 in Algorithm~\ref{alg:rta})

\subsubsection{Writing refinement}

With factual integrity secured, we shift focus to stylistic enhancements to make the trace more coherent, readable, and pedagogically engaging without compromising accuracy. Starting from the audited trace $r$ (not the initial $r_0$), $G$ iteratively polishes it (see Algorithm~\ref{alg:stylepolish}). It includes optimization of target coherence and readability, quantified via an average reward score. Crucially, we monitor factuality at every step; if the new F1 score drops below $F1_{\text{optimal}}$ or the reward decreases, refinement halts immediately to prevent degradation. Otherwise, it continues until the reward stabilizes (difference below $\epsilon$ for two consecutive iterations).

For feedback refinement (Stage 3 in Algorithm~\ref{alg:rta}), we apply only the factual auditing process to the initial output $y_0$, producing a refined feedback $\hat{f}$ with its own optimal F1 score. Stylistic polishing is omitted here, as these elements are inherited from the aligned reasoning trace in later stages.

This refinement strategy produces structured, verifiable traces that not only align with domain constraints but also serve as high-quality supervision, enabling models to deliver dependable diagnostics in constraint-heavy tasks.

\subsection{Two-Stage Model Alignment}

Building on the refined traces, R$^2$tA employs a sequential alignment process to instill reasoning and feedback capabilities into the model. It leverages SFT and DPO for efficient, reward-model-free optimization. In Stage 2 (Algorithm~\ref{alg:rta}), We first apply SFT to calibrate the base model $\pi_0$ on dataset $\mathcal{R}$, optimizing via maximum likelihood to produce coherent, domain-aligned reasoning $r_{\text{sft}}$. Next, we construct preference pairs $\mathcal{P}_r$ using refined traces $\hat{r}$ as chosen responses and suboptimal generations ($r_{\text{sft}}$) as rejected, then employ DPO to align $\pi_r$ with conceptual preferences, ensuring outputs condition faithfully on rationales without reward modeling. This yields a reasoning-aligned model $\pi_r$ with enhanced fidelity for complex tasks.\par

R$^2$tA then conditions feedback on these aligned traces for output fidelity (Stage 4 in Algorithm~\ref{alg:rta}). We first generate aligned reasoning $r^*$ from $\pi_r$ for each $x$, build training dataset $\mathcal{F}_{\text{train}}$ pairing inputs with $r^*$ and refined feedback $\hat{f}$, then apply SFT from $\pi_r$ via maximum likelihood to calibrate coherent feedback $f_{\text{sft}}$. Next, we construct preference pairs $\mathcal{P}_f$ using $\hat{f}$ as chosen responses and suboptimal generations as rejected, then employ Direct Preference Optimization (DPO) to align $\pi_f$ with conceptual preferences, ensuring outputs condition on faithful rationales without reward modeling. This yields a feedback-aligned model $\pi_f$ that produces actionable, mistake-level outputs grounded in domain-aligned reasoning.

By iteratively refining and aligning through these stages, R$^2$tA bridges the gap between raw LLM generations and expert-level performance, making it a robust framework.

\begin{algorithm}[t]
\small
\caption{Iterative Factual Audit}
\label{alg:factaudit}
\begin{algorithmic}[1]
\STATE \textbf{Input:} Guide LLM $G$, initial response $resp$, max iterations $M$, threshold $\epsilon$
\STATE \textbf{Output:} Refined response $\hat{resp}$, optimal F1 score $F1_{optimal}$
\STATE $\hat{resp} \leftarrow resp$, $F1_{prev} \leftarrow 0$, $sat_{count} \leftarrow 0$
\FOR{$i = 1$ to $M$}
  \STATE $\hat{resp} \leftarrow \textsc{AuditFactuality}(G, \hat{resp})$ \textit{// Detect/remove hallucinations, insert missed explanations}
  \STATE $F1_{curr} \leftarrow \textsc{ComputeF1}(\hat{resp})$ \textit{// Against schema-specific rubrics}
  \IF{$|F1_{curr} - F1_{prev}| < \epsilon$}
    \STATE $sat_{count} \leftarrow sat_{count} + 1$
    \IF{$sat_{count} \geq 2$}
      \STATE \textbf{break}
    \ENDIF
  \ELSE
    \STATE $sat_{count} \leftarrow 0$
  \ENDIF
  \STATE $F1_{prev} \leftarrow F1_{curr}$
\ENDFOR
\STATE $F1_{optimal} \leftarrow F1_{prev}$
\STATE \textbf{return} $\hat{resp}, F1_{optimal}$
\end{algorithmic}
\end{algorithm}

\begin{algorithm}[t]
\small
\caption{Iterative Style Polish}
\label{alg:stylepolish}
\begin{algorithmic}[1]
\STATE \textbf{Input:} Guide LLM $G$, initial response $resp$, initial F1 score $F1_{init}$, max iterations $M$
\STATE \textbf{Output:} Refined response $\hat{resp}$
\STATE $\hat{resp} \leftarrow resp$, $reward_{prev} \leftarrow 0$
\FOR{$i = 1$ to $M$}
  \STATE $resp' \leftarrow \textsc{PolishStyle}(G, \hat{resp})$ \textit{// Optimize coherence, readability; stop early if F1 drops below $F1_{init}$ or reward decreases}
  \STATE $reward_{curr} \leftarrow \textsc{ComputeAvgReward}(resp')$ \textit{// Average coherence/readability scores}
  \STATE $F1_{fact} \leftarrow \textsc{ComputeF1}(resp')$
  \IF{$F1_{fact} \geq F1_{init}$ \AND $reward_{curr} > reward_{prev}$}
    \STATE $\hat{resp} \leftarrow resp'$
    \STATE $reward_{prev} \leftarrow reward_{curr}$
  \ENDIF
\ENDFOR
\STATE \textbf{return} $\hat{resp}$
\end{algorithmic}
\end{algorithm}

\begin{algorithm}[t]
\small
\caption{Reason-Refine-then-Align (R$^2$tA)}
\label{alg:rta}
\begin{algorithmic}[1]
\STATE \textbf{Input:} Prompt dataset $\mathcal{D}x$ (prompt + task input), base LLM $\pi_0$, guide LLM $G$, SFT epochs $N_{sft}$, DPO steps $N_{\text{dpo}}$, LoRA configs, max iterations $M$, threshold $\epsilon$
\STATE \textbf{Output:} Reasoning-aligned model $\pi_r$, feedback-aligned model $\pi_f$ \\
\textbf{Stage 1: Reasoning refinement (from $\pi_0$)}
\FORALL{$x \in \mathcal{D}x$}
\STATE $(r_0, y_0) \leftarrow \pi_0(x)$
\STATE $(r, F1_{optimal}) \leftarrow \textsc{IterativeFactualAudit}(G, r_0, M, \epsilon)$ \textit{// Factual auditing (iterative until F1 stabilizes)}
\STATE $\hat{r} \leftarrow \textsc{IterativeStylePolish}(G, r, F1_{optimal}, M, \epsilon)$ \textit{// Stylistic polishing (iterative with factuality checkpoint)}
\STATE $\mathcal{R} \leftarrow \mathcal{R} \cup {(x, \hat{r})}$
\ENDFOR \\
\textbf{Stage 2: Reasoning alignment (input = prompt + task input)}
\STATE $\pi_r \leftarrow \textsc{SFT}(\pi_0,\mathcal{R},N_{\text{sft}})$
\STATE Build $\mathcal{P}r = {(\hat{r}, r_{\text{sft}} \mid r_{\text{sft}}=\pi_r(x), x\in\mathcal{D}_x})$
\STATE $\pi_r \leftarrow \textsc{DPO}(\pi_r,\mathcal{P}r,N_{\text{dpo}})$ \\
\textbf{Stage 3: Feedback refinement}
\FORALL{$x \in \mathcal{D}_x$}
\STATE $f_0 \leftarrow y_0$
\STATE $(\hat{f}, F1_{optimal}) \leftarrow \textsc{IterativeFactualAudit}(G, f_0, M, \epsilon)$ \textit{// Factual auditing only (iterative until F1 stabilizes)}
\STATE $\mathcal{F} \leftarrow \mathcal{F} \cup {(x, \hat{f})}$
\ENDFOR \\
\textbf{Stage 4: Feedback alignment (input = prompt + task input + aligned reasoning)}
\FORALL{$x \in \mathcal{D}x$}
\STATE $r^* \leftarrow \pi_r(x)$
\STATE fetch $\hat{f}$ for $x$ from $\mathcal{F}$
\STATE $\mathcal{F}{\text{train}} \leftarrow \mathcal{F}{\text{train}} \cup {(x, r^*, \hat{f})}$
\ENDFOR
\STATE $\pi_f \leftarrow \textsc{SFT}(\pi_r,\mathcal{F}{\text{train}},N_{\text{sft}})$
\STATE Build $\mathcal{P}f = {(\hat{f}, f_{\text{sft}} \mid f_{\text{sft}}=\pi_f(x, r^*), x\in\mathcal{D}_x}$)
\STATE $\pi_f \leftarrow \textsc{DPO}(\pi_f,\mathcal{P}f,N_{{dpo}})$
\STATE \textbf{return} $\pi_r, \pi_f$
\end{algorithmic}
\end{algorithm}

\section{Experiments}

\subsection{Dataset curation}
Due to a lack of existing benchmarks, we created a dataset by adapting three EERD specifications from a database systems textbook~\citet{ElmasriNavathe2016}\footnote{This dataset generation process is explained in supplementary material}. For training, we adapted the specifications from the company (Fig.~3.2), university (Fig.~4.9), and small airport (Fig.~4.12) database EERD schemas. For each schema, we intentionally induced mistakes across 11 categories\footnote{Mistake spans across 11 categories- key attribute, total participation, ternary relationships, specialization/union, cardinalities, relationship participants, Attribute types, Attributes, entity types, invalid relationships, relationship types}. -creating 150 mistaken EERD variants per schema. While inducing mistakes, we ensured that single and multiple mistakes coexist in the same EERD by randomly sampling from the categories, yielding 450 variants across the three schemas for training. For evaluation, we held out a completely new schema (Hospital database) that was unavailable during training and generated another 150 mistaken EERD variants following the same protocol.

\subsection{Rubric Design}
Motivated by prior work on LLM-driven feedback for EERD evaluation~\cite{riazi2025llm}, which demonstrates how structured rubrics enhance LLM accuracy in assessing conceptual designs without relying on a single correct solution, we incorporate schema-specific rubrics to guide refinement and inject domain-aligned conceptual preferences. For each EERD schema in our dataset, we design rubrics comprising two key components: (1) problem statements that contextualize elements (\ie relationships, etc) within the domain, drawing from textbook descriptions~\cite{ElmasriNavathe2016}; and (2) evaluation criteria that define objective assessment points for constraints like participation, cardinalities, etc. These rubrics are integrated into guide LLM prompts\footnote{Rubrics for a single schema and detailed prompt for the refinement is available in the supplementary material} during refinement (Stages 1 and 3 in Algorithm~\ref{alg:rta}), enabling fact verification against domain requirements and correction of hallucinations while preserving conceptual integrity. This approach ensures refined traces reflect real-world modeling implications.

\subsection{Baselines}

To benchmark R$^2$tA, we compare it against six baselines on the EERD task. R$^2$tA uses a base model with reasoning LoRA (SFT + DPO) and feedback LoRA (SFT + DPO).

\begin{itemize}
    \item \textbf{DeepSeek Base (DSB)}: Zero-shot with DeepSeek-R1-Distill-Qwen-32B ($\pi_0$).
    \item \textbf{GPT-4o}: Direct inference with GPT-4o.
    \item \textbf{Base + Feedback LoRA (SFT) (B+Fb-SFT)}: DeepSeek Base fine-tuned via SFT on feedback.
    \item \textbf{Base + Feedback LoRA (SFT + DPO) (B+Fb-SFT+DPO)}: As above, plus DPO.
    \item \textbf{Base + Reasoning LoRA (SFT + DPO) + Feedback LoRA (SFT + DPO) (R$^2$tA)}: Full R$^2$tA.
    \item \textbf{Base + Reasoning LoRA (SFT + DPO) + Feedback LoRA (SFT) (R$^2$tA-noFbDPO)}: R$^2$tA without DPO on feedback.
\end{itemize}

These ablate R$^2$tA's components for fair insight. All train on the same dataset. We test on held-out Hospital schema.

\subsection{Evaluation Metrics}
We use GPT-4o~\cite{hurst2024gpt} to evaluate feedback against ground-truth EERD mistakes at the per-mistake level. The evaluation prompt\footnote{Evaluation prompt is available in the supplementary note} includes ground-truth types, schema rubrics, correct EERD, mistaken EERD, and  output. It checks detection per mistake (even partial), extracts phrases for detections/misses, identifies false positives with rationale, and suggests ideal explanations. We compute TP (correctly detected mistakes), FN (missed ground-truth mistakes), and FP (hallucinated mistakes); precision as TP / (TP + FP) for accuracy; recall as TP / (TP + FN) for coverage; and F1 as 2 × (precision × recall) / (precision + recall) for balance. Metrics average across examples, with category breakdowns.

\subsection{Implementation Details}
We use GPT-4o~\cite{hurst2024gpt} as a guide LLM $G$ for refinement. Temperature set to 0 for reproducibility. Max iterations are $M=5$. Convergence threshold is $\epsilon=0.05$. This setup ensures efficient audits.
For alignment, we adapt DeepSeek-R1-Distill-Qwen-32B\footnote{https://huggingface.co/deepseek-ai/DeepSeek-R1-Distill-Qwen-32B} as base LLM $\pi_0$. We apply 4-bit quantization for efficiency. SFT stages run for 1 epoch. DPO stages take 50 steps. Refined traces serve as chosen preferences. Post-SFT outputs act as rejected.\footnote{Full hyperparameters appear in the supplementary material.}

\section{Results}

\subsection {Enhanced Reasoning Significantly Improves Detection in Complex Mistake Categories}

To evaluate the effectiveness of R$^2$tA, we particularly examined mistake categories demanding substantial reasoning, such as Specialization or Union, Ternary Relationships, and Relationship Types (see Table \ref{tab:model_scores}). Our method consistently outperformed all baselines across these reasoning-intensive categories. Specifically, in detecting errors within Ternary Relationships, R$^2$tA achieved an F1 of 89, substantially exceeding strong baselines like GPT-4o (82) and the non-reasoning-enhanced DeepSeek Base (68). Similarly, in the structurally nuanced Specialization or Union category, R$^2$tA’s iterative refinement and alignment led to an F1 improvement (64) over GPT-4o (60) and dramatically improved over DeepSeek Base (24). These results highlight R$^2$tA’s ability to rectify complex conceptual mistakes through refined intermediate reasoning, leading to more accurate and pedagogically actionable feedback.

\subsection {Ablation Study: Impact of Two-Stage Alignment (SFT+DPO)}
We further analyzed the impact of our two-stage alignment approach, combining SFT with DPO, through targeted ablations (see Table \ref{tab:model_scores}). The complete R$^2$tA framework (Reasoning LoRA (SFT+DPO) + Feedback LoRA (SFT+DPO)) achieved the highest overall average F1 (76), significantly outperforming the baseline without feedback DPO alignment (R$^2$tA-noFbDPO: F1 = 74) and baseline models trained with SFT-only (B+Fb-SFT: F1 = 42). This improvement demonstrates that aligning both intermediate reasoning and final feedback explicitly with conceptual preferences via DPO is essential to achieving reliable performance. The integration of DPO notably improved precision, as evidenced by R$^2$tA’s higher average precision (79) compared to R$^2$tA-noFbDPO (77), underscoring the critical role of preference-based alignment in mitigating false positives.

\subsection{Human Evaluation}
We use expert evaluation to compare the quality of feedback generated by the fine-tuned Deepseek model, R$^2$tA, and GPT-4o. We evaluated 60 feedback samples generated for student submissions (no synthetic induced errors) in an unseen environment, not used during the fine-tuning or testing of the Deepseek models.

Our primary evaluation metric is the F1 score -- which excludes true negatives since they do not impact either precision or recall. Through expert review, we found that R$^2$tA’s feedback was more accepting of correct design choices, especially in the categories of Specialization and Generalization. In these cases, students introduced correct specialization or generalization that was not explicitly mentioned in the problem statement. While GPT-4o generally avoided commenting on these new, correct usages, R$^2$tA  explicitly recognized and affirmed them. An example output from both models is shown below.

R$^2$tA: \textit{
"The Nurse entity incorrectly uses Physician\_ID as a key attribute instead of having its own unique Nurse\_ID. The Medication entity is incorrectly modeled as a weak entity with partial keys; it should be part of the Payables union. The `Executes` relationship is correctly modeled as a ternary relationship with appropriate cardinality and total participation for OrderInstructions. The Payables union correctly includes Medication, Room, and Instruction as supercategories. The feedback should address these points clearly and concisely, ensuring each entity and relationship is evaluated against the problem statement."}

GPT-4o: \textit{
"The submission correctly identifies `Executes` as a ternary relationship among Nurse, OrderInstructions, and Patient, which aligns with the problem statement. The cardinality of all participating entities is correctly identified as `N`, which matches the problem statement`s description of the relationship. The attribute `Status` is correctly included as an attribute of the `Executes` relationship. However, there is an error in the entity type for `OrderInstructions`, which should be `Instruction` as per the problem statement. Additionally, `Nurse` is incorrectly assigned a key attribute `Physician\_ID`, which should be `Nurse\_ID`. The entity `Medication` is correctly identified as a weak entity with partial keys `Type` and `Amount`. The total participation of `OrderInstructions` in the `Executes` relationship is correctly identified."
}

Among all the feedback categories, only Cardinality, Keys, and Total Participation had enough student mistakes to allow for a meaningful quantitative comparison. For Cardinality, R$^2$tA achieved a precision, recall, and F1 score of 89, 80, and 84 (reported as 89/80/84), while GPT-4o achieved 67/54/60. For Keys, R$^2$tA scored 83/50/62 compared to GPT-4o’s 60/43/50. For Total Participation, R$^2$tA performed at 69/47/56, while GPT-4o had much lower scores at 50/11/18. In this last category, GPT-4o generated many false negatives, incorrectly acknowledging correct total participation when it was actually missing from student submissions. 

Overall, we observed that the fine-tuned Deepseek model using R$^2$tA generated far fewer unsupported statements and provided more accurate feedback compared to GPT-4o.

\subsection {Discussion}

Our results demonstrate that output-only fine-tuning degrades the faithfulness of intermediate reasoning traces, consistent with prior findings~\cite{lobo2025impact}.

In Table \ref{tab:model_scores}, feedback-only baselines (B+Fb-SFT and B+Fb-SFT+DPO) exhibit high variance across concept-heavy categories: they achieve competitive F1 on Ternary Relationships (76 and 82, respectively) but collapse on Relationship Types (down to 15 for B+Fb-SFT+DPO) and Specialization/Union (36 for B+Fb-SFT+DPO). This indicates that output-focused tuning lacks structural guidance to preserve causal, mistake-level reasoning, confirming that raw fine-tuning without reasoning refinement fails to generalize.

Future work should consider applying this structuralist refinement method across broader educational contexts, investigate explicit causal modeling for deeper interpretability. First, extend structural auditing beyond ERDs to other graph-structured domains (\ie  UML, knowledge graphs) and add explicit causal checks linking each reasoning step to schema constraints. Second, enhance evaluation fidelity: supplement or replace LLM-based graders with programmatic checkers.

\begin{table*}[!ht]
\centering
\small 
\caption{Precision, Recall, and F1 Scores across Baselines and Mistake Categories. Entries are formatted as Precision/Recall/F1, with bold indicating the highest F1 score(s) per category.}
\label{tab:model_scores}
\resizebox{\textwidth}{!}{
\begin{tabular}{|l|c|c|c|c|c|c|}
\hline
\textbf{Mistake Category} & \textbf{B+Fb-SFT+DPO} & \textbf{B+Fb-SFT} & \textbf{Full R$^2$tA} & \textbf{R$^2$tA-noFbDPO} & \textbf{DSB} & \textbf{GPT-4o} \\
\hline
Keys & 56/58/57 & 38/56/45 & 78/69/\textbf{73} & 65/75/70 & 69/56/62 & 71/65/68 \\
\hline
Total Participation & 80/75/77 & 61/64/62 & 85/73/\textbf{79} & 81/72/76 & 76/55/64 & 79/71/74 \\
\hline
Ternary Relationships & 81/83/82 & 70/82/76 & 94/85/\textbf{89} & 82/78/80 & 78/60/68 & 83/81/82 \\
\hline
Specialization/Union & 35/37/36 & 19/23/21 & 62/67/64 & 68/74/\textbf{71} & 30/20/24 & 59/61/60 \\
\hline
Cardinalities & 81/83/82 & 52/54/53 & 83/78/80 & 84/76/80 & 78/63/70 & 81/88/\textbf{84} \\
\hline
Participants & 74/80/77 & 17/18/18 & 82/77/\textbf{79} & 80/75/77 & 62/58/60 & 75/77/76 \\
\hline
Attribute Types & 74/68/\textbf{71} & 27/22/24 & 56/68/61 & 66/59/62 & 58/55/56 & 61/63/62 \\
\hline
Attributes & 72/76/74& 26/22/24 & 78/73/\textbf{75} & 73/71/72 & 44/42/43 & 69/75/72 \\
\hline
Entity Types & 83/78/\textbf{80} & 35/38/36 & 83/73/78 & 82/68/74 & 81/66/73 & 82/75/78 \\
\hline
Invalid Relationships & 61/63/62 & 38/36/37 & 83/64/72 & 83/62/71 & 72/58/64 & 78/81/\textbf{79} \\
\hline
Relationship Types & 12/19/15 & 68/63/65 & 85/82/\textbf{83} & 78/76/77 & 74/60/66 & 77/74/75 \\
\hline\hline
\textbf{Reasoning Enhanced} & \textcolor{nored}{\ding{55}} & \textcolor{nored}{\ding{55}} & \textcolor{yesgreen}{\ding{51}} & \textcolor{yesgreen}{\ding{51}} & \textcolor{nored}{\ding{55}} & \textcolor{nored}{\ding{55}} \\
\hline\hline
\hline\hline \textbf{Average} & 64/65/65 & 41/43/42& 79/74/\textbf{76} & 77/71/74 & 66/54/59 & 74/74/74 \\

\hline
\end{tabular}}
\end{table*}

\section{Related work}

Our work builds on recent advancements in adapting LLMs for specialized reasoning tasks, particularly through refinement of generated rationales and alignment techniques. \par
Existing methods for \textbf{eliciting and distilling (CoT) reasoning} have laid the groundwork for improving model transparency and performance on complex problems. For instance,~\citet{wei2022chain} introduced CoT prompting to enable step-by-step reasoning in LLMs, while subsequent distillation approaches like SCOTT~\cite{wang2023scott} and Symbolic CoT Distillation~\cite{li2023symbolic} use faithful knowledge distillation to transfer CoT capabilities from large teachers to smaller students, ensuring self-consistency and enabling smaller models to "think" effectively. Similarly,~\citet{li-etal-2024-mode} decouples CoT for complex tasks across model sizes, and Keypoint-based Progressive CoT Distillation~\cite{feng2024keypoint} identifies critical reasoning points for progressive refinement, addressing efficiency in transfer. However, these methods often rely on high-quality initial rationales without explicit auditing for errors, which R$^2$tA addresses by incorporating LLM-guided step-wise correction prior to alignment.\par
\textbf{Self-correction and refinement mechanisms in LLMs}: Recent efforts exploit LLMs' ability to verify and iteratively improve outputs, often without external supervision. Self-Refine~\cite{madaan2023self} employs self-feedback loops to refine generations across domains, while CRITIC~\cite{gou2023critic} allows LLMs to validate and amend outputs using tools. Training via reinforcement learning, as in SCoRe~\cite{gou2023critic}, enhances self-correction for multi-turn scenarios, and studies~\cite {huang2023large,kamoi2024can} reveal that LLMs cannot reliably self-correct general reasoning without structured guidance, highlighting inconsistencies in complex domains. Embedding self-correction as in CoSC~\cite{gao2024embedding} focuses on iterative refinement for mathematical tasks. Unlike these intrinsic approaches, which risk propagating biases or missing nuances, R$^2$tA uses a separate guide LLM with domain-specific rubrics for auditing, ensuring comprehensive error coverage before fine-tuning.\par
\textbf{Fine-tuning and alignment techniques}: SFT~\cite{radford2018improving} followed by preference optimization has become standard for aligning LLMs with human or task preferences. DPO~\cite{rafailov2023direct} optimizes models directly on preferences without reward models. However, these processes can degrade CoT faithfulness; for example,~\citet{lanham2023measuring} observe inverse scaling where RLHF-finetuned models~\cite{ouyang2022training} exhibit less faithful reasoning as size increases, with larger models showing more post-hoc rationalization. Similarly,~\citet{bao2025likely} finds that SFT introduces spurious features and weakens causal structures in CoT, while RLHF~\cite{ouyang2022training} (via DPO~\cite{rafailov2023direct}) further reduces the influence of reasoning steps on final outputs, leading to increased hallucinations and unfaithfulness.~\citet{bentham2024chain} reports similar degradation in DPO-finetuned models, with inverse scaling in faithfulness starting at smaller parameter sizes. R$^2$tA differentiates by decoupling refinement from alignment, using audited traces for SFT to calibrate reasoning, followed by DPO for output conditioning, preserving faithfulness. \par
\textbf{Verification and auditing of LLM outputs}: To mitigate hallucinations, auditing frameworks verify outputs for consistency and actuality. LLMAuditor ~\cite{amirizaniani2024llmauditor} audits LLMs for vulnerabilities using multi-agent approaches, while WildHallucinations~\cite{zhao2024wildhallucinations} evaluates long-form factuality across domains. PRISM ~\cite{azzopardi2024prism} audits biases through inquiry-based methods and Weakly supervised detection ~\cite{rateike2023weakly} identifies hallucinations via activation patterns. While effective for post-hoc checks, these often overlook domain-specific nuances. R$^2$tA integrates structured auditing with rubrics during refinement, detecting false positives/negatives to yield trustworthy supervision for alignment. \par
\textbf{LLM applications in education and structured tasks}: LLMs are increasingly applied to educational feedback, especially in domains requiring structured reasoning like database design. LLM-driven systems provide targeted feedback for conceptual designs in database courses ~\cite{riazi2025llm}, CoddLLM ~\cite{zhang2025coddllm} empowers LLMs for data analytics tasks, ERBench ~\cite{oh2024erbench} benchmarks LLMs on entity-relationship questions verifiable via databases, and multi-agent debates enhance requirements elicitation in education ~\cite{oriol2025multi}. Surveys on LLMs in education ~\cite{xu2024large} emphasize reliability for structured tasks, but prompt-based methods often hallucinate in EERD evaluation ~\cite{riazi2025llm}. R$^2$tA advances this by curating EERD variants with induced errors and refining rationales for faithful, actionable feedback.\par
In contrast to prior works, which either focus on distillation without correction or alignment with abundant labels, R$^2$tA uniquely decouples LLM-guided auditing from two-stage SFT-DPO alignment, enabling label-efficient adaptation for structured reasoning in education, with demonstrated improvements on EERD evaluation.

\section{Conclusion}
Traditional fine-tuning methods fall short in preserving faithful Chain-of-Thought reasoning, particularly in complex, schema-driven tasks like EERD analysis, as evidenced by high variance and performance collapses in concept-heavy categories. Our iterative reasoning refinement framework addresses this by decoupling step-level factual auditing from structural polishing, yielding consistent gains in multi-step relational detection across scales. These results underscore that structure-aware refinement is key to generalizable, mistake-resilient reasoning, paving the way for broader applications in graph-structured domains and enhanced evaluation protocols. Ultimately, this approach advances towards more reliable AI systems capable of causal, transparent inference under real-world constraints.

\bibliography{aaai2026}

\appendix
\onecolumn

\section{Additional experiment}
To test the generalizability and robustness of reasoning enhancement, we compared R$^2$tA’s performance on a reduced model size more computationally efficient 7B variant (see Table  \ref{tab:model_scores7b}). R$^2$tA demonstrated consistent performance gains even on the smaller 7B model, achieving an average F1 of 55, outperforming both DeepSeek Base (F1 = 46) and the no-feedback-DPO variant (F1 = 51). Although smaller models inherently show performance limitations compared to larger counterparts, the notable relative improvement highlights the efficacy and versatility of R$^2$tA’s iterative reasoning refinement across model scales. This finding suggests potential practical deployments where computational resources are constrained without compromising significant improvements in structured reasoning tasks.
\begin{table*}[!htbp]
\centering
\footnotesize 
\caption{Precision, Recall, and F1 Scores across Models by Mistake Categories for 7B Parameter Models ( DeepSeek-R1-Distill-Qwen-7B). Entries are formatted as Precision/Recall/F1. Bold indicates the highest F1 score(s) in each category.}
\label{tab:model_scores7b}
\begin{tabular}{|l|c|c|c|c|}
\hline
\textbf{Mistake Category} & \textbf{Full R$^2$tA} & \textbf{R$^2$tA-noFbDPO} & \textbf{DSB} & \textbf{GPT-4o} \\
\hline
Keys &  61/59/60 & 56/49/52 & 49/51/50 & 71/65/\textbf{68} \\
\hline
Total Participation & 65/68/66 & 64/62/63 & 52/54/53 & 79/71/\textbf{74} \\
\hline
Ternary Relationships  & 63/65/64 & 61/58/59 & 57/60/58 & 83/81/\textbf{82} \\

\hline
specialization or Union  & 42/37/39 & 38/32/35 & 26/22/238 & 59/61/\textbf{60} \\

\hline
Cardinalities & 62/68/65 & 58/66/62& 52/58/55 & 81/88/\textbf{84} \\

\hline
Participants & 46/50/47& 40/37/38 & 39/41/40 & 75/77/\textbf{76} \\

\hline
Attribute Types & 47/41/44 & 46/51/48 & 48/45/46 & 61/63/\textbf{62} \\

\hline
Attributes & 59/63/60 & 52/55/53 & 35/42/38 & 69/75/\textbf{72} \\

\hline
Entity Types & 69/64/66 & 62/59/60 & 61/57/59 & 82/75/\textbf{78} \\
\hline
Invalid Relationships & 48/47/48 & 43/52/47 & 39/45/417 & 78/81/\textbf{80} \\
\hline
Relationship Types & 49/52/50\ & 42/48/45 & 44/46/45 & 77/74/\textbf{75} \\
\hline\hline
\textbf{Reasoning Enhanced} & \textcolor{yesgreen}{\ding{51}} & \textcolor{yesgreen}{\ding{51}} & \textcolor{nored}{\ding{55}} & \textcolor{nored}{\ding{55}} \\
\hline\hline
\hline\hline \textbf{Average} & 55/56/55 & 51/52/51 & 45/47/46 & \textbf{74/73/73} \\

\hline
\end{tabular}
\end{table*}

\section{Rubrics Design}

Below we demonstrate how we designed rubrics for a specific  schema, which are subsequently used for retrieving relevant evaluation statements. These rubrics serve as human-validated conceptual preferences that guide LLMs in evaluating submitted database designs against established best practices and theoretical foundations. 

\begin{figure}[H]
    \centering
    \includegraphics[width=0.8\textwidth]{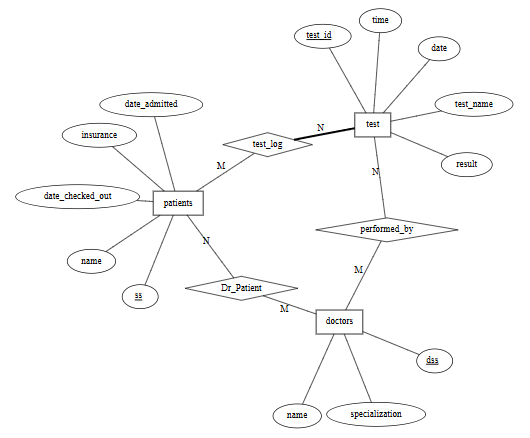}
    \caption{The  schema used for following rubric development}
    \label{fig:erd_schema}
\end{figure}

\vspace{1em}

\begin{lstlisting}[language=json]
{
    "problem-statements": [
        {
            "description": "The patients entity captures essential information including ss, name, insurance, date_admitted and date_checked_out.",
            "rubrics": [
                "patients is a strong entity, uniquely identified by the attribute ss.",
                "name, insurance, date_admitted and date_checked_out are modeled as simple attributes."
                 ],
            "questions": [
                "How would the model handle patients with multiple admissions?",
                "Should insurance be modeled as a composite attribute to capture policy number, coverage details etc?"
               
            ]
        },
        {
            "description": "The doctors entity identified by dss and characterized by name and specialization.",
            "rubrics": [
                "doctors is a strong entity with dss as the primary key.",
                "Specialization and name are modeled as a simple attribute."
               
            ],
            "questions": [
               
                "Should specialization be a multivalued attribute to represent doctors with multiple specialties?"
               
            ]
        },
        {
            "description": "The test entity captures the detailed description of test results.",
            "rubrics": [
                "test is a strong entity identified by test_id.",
                "test includes other attributes: test_name, date, time, and result."
               
            ],
            "questions": [
                "Should test_id be globally unique or uniqueness limited to patient context?",
                "Why are date and time separate attributes instead of a single datetime attribute?"
               
            ]
        },
        {
            "description": "test_log records which tests belong to which patients. ",
            "rubrics": [
                "test_log is a binary many to many (M:N) relationship between patients and test.",
                "test participates totally in test_log.",
                "patients participation is partial, a patient may have zero tests."
            ],
            "questions": [
              
                "Would test_log still be necessary if tests were modeled as a weak entity under patients?",
                "Why use many to many (M:N) relationship for test_log instead of 1:N (one test per patient)?"
            ]
        },
        {
            "description": "tests are conducted by doctors through the performed_by relationship, allowing multiple doctors to collaborate.",
            "rubrics": [
                "performed_by is a binary M:N relationship between doctors and test.",
                "Each test can be performed by multiple doctors.",
                "Each doctor can perform multiple tests."
            ],
            "questions": [
                "Should performed_by distinguish between the primary doctor and assisting doctors?",
                "Should test participate totally in PERFORMED_BY?"
            ]
        },
        {
            "description": "patients and doctors have a many-to-many relationship through Dr_Patient.",
            "rubrics": [
                "Dr\_Patient is a binary many to many (M:N) relationship between doctors and patients.",
                "Each patient can be treated by multiple doctors.",
                "Each doctor can treat multiple patients."
            ],
            "questions": [
                "Why is Dr_Patient modeled as M:N rather than 1:N",
                "What if all doctor treatments are already captured by performed_by?"
            ]
        }
    ]
}
\end{lstlisting}

\section{Progressive Error Injection Demonstrates Effectiveness }

To demonstrate the effectiveness of our approach, we gradually insert mistakes into a specific relationship in the aforementioned schema (\textbf{test\_log relation}) and evaluate performance under increasing error complexity.

\textbf{Mistake 0: No mistake}

\begin{figure}[H]
    \centering
    \includegraphics[width=0.39\textwidth]{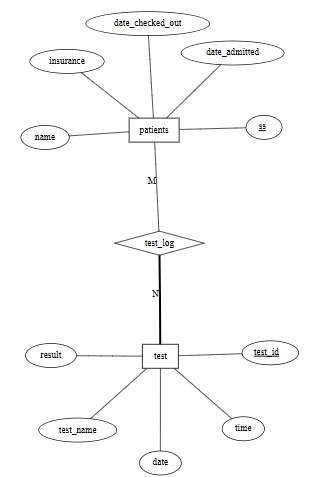}
    \caption{The schema}
    \label{fig:erd_schema}
\end{figure}

\textbf{ASSESSMENT}: Both models positively provide feedback. \\

Gpt-4o: The submission correctly identifies `patients` as a strong entity with `ss` as the key attribute, and `test` as a strong entity with `test\_id` as the key attribute. The `test\_log` relationship is correctly modeled as a many-to-many (M:N) relationship between `patients` and `test`. The total participation of `test` in `test\_log` is correctly identified, while `patients` participation is partial, which aligns with the problem statement.
\\

Full R$^2$tA:
The ER model submission accurately represents the entities and relationships as described in the problem statement.
 Both patients and test are correctly modeled as strong entities, with `ss` and `test\_id` serving as their respective key attributes.
 The `test\_log` relationship is appropriately defined as a many-to-many (M:N) relationship, reflecting the non-identifying nature of the connection between the two strong entities. 
 Total participation is correctly assigned to the `test` entity, ensuring that every test is associated with at least one patient, while `patients` participation is optional, allowing for patients without any tests.
 All specified attributes are present in the submission, and no additional feedback is required beyond confirming the accuracy of the model against the provided criteria.

\textbf{Mistake 1: remove the total participation constraint}

\begin{figure}[H]
    \centering
    \includegraphics[width=0.39\textwidth]{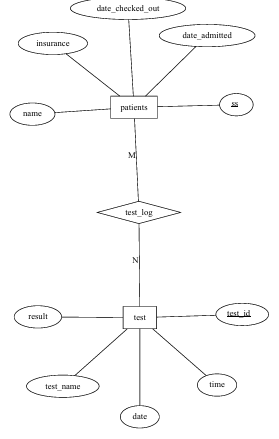}
    \caption{The schema}
    \label{fig:erd_schema}
\end{figure}

\textbf{ASSESSMENT}: Both models successfully identify the single error in the schema. This detection is expected given the severity of the mistake.\\

\noindent GPT-4o: The submission correctly identifies `patients` and `test` as strong entities. The `test\_log` relationship is correctly modeled as a many-to-many (M:N) \correct{relationship between `patients` and `test`. However, the submission does not specify the total participation of `test` in the `test\_log` relationship, which is required according to the problem statement.} Additionally, the partial participation of `patients` in `test\_log` is not explicitly mentioned in the submission.
\\

\noindent Full R$^2$tA:The entities `patients` and `test` are correctly modeled as strong entities with appropriate key attributes (`ss` for patients and `test\_id` for tests). The relationship `test\_log` is correctly identified as a many-to-many (M:N) relationship between `patients` and `test`. \correct{However, the submission does not specify total participation for `test` in the `test\_log` relationship,} which is a critical point according to the problem statement. Additionally, the separation of `date` and `time` into different attributes instead of a single `datetime` is noted in the problem statement, and the uniqueness of `test\_id` could be worth considering in feedback.

\noindent\textbf{Mistake 2: remove the total participation constraint+key attribute in test entity}
\begin{figure}[H]
    \centering
    \includegraphics[width=0.39\textwidth]{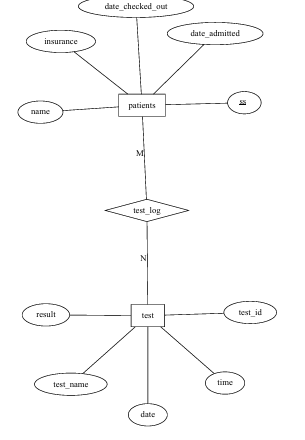}
    \caption{The  schema used for the following rubric development}
    \label{fig:erd_schema}
\end{figure}
\textbf{ASSESSMENT}: Although GPT-4o hallucinates regarding the presence of the key attribute, we observe another interesting pattern: GPT-4o struggles to identify the relationship type despite correctly recognizing both entities as strong. This provides clear evidence for why reasoning-level verification is essential. In contrast, our model derives the relationship type based on entity classification (when two entities are strong, the relationship is non-identifying). This demonstrates that if LLMs can be trained to reason conceptually, they can correctly infer such patterns.\\

Gpt-4o: The submission correctly identifies `patients` as a strong entity with `ss` as the key attribute, aligning with the problem statement. \incorrect{The `test` entity is also correctly identified as a strong entity with `test\_id` as the key attribute.} The `test\_log` relationship is correctly modeled as a many-to-many (M:N) relationship between `patients` and `test`. However, the submission does not specify whether the `test\_log` relationship is identifying or non-identifying, which is necessary for complete verification. \correct{Additionally, the total participation of `test` in `test\_log` is not explicitly mentioned in the submission}, which is required according to the problem statement.
\\

Full R$^2$tA: The feedback evaluates the submitted ERD against the problem statement, focusing on entity types, relationship types, and participation. The `patients` entity is correctly modeled as a strong entity with `ss` as a key attribute. \correct{However, the `test` entity should be a strong entity, but `test\_id` is not marked as a key attribute in the submission.} The `test\_log` relationship is correctly identified as a non-identifying many-to-many (M:N) relationship. Participation is incorrect; \correct{`test` must have total participation, while `patients` should be partial.} The feedback highlights these issues and their implications for the model`s accuracy.
\vspace{1em}
\hrule
\vspace{1em}
B+FbSFT+DPO (Baseline w/o reasoning enhancement):

The submission accurately models the entities and relationships as described in the problem statement.
 \incorrect{The "patients" entity is correctly identified as a strong entity with "ss" as the key attribute, and all other attributes are appropriately modeled as simple attributes. }
 \incorrect{Similarly, the "test" entity is correctly identified as a strong entity with "test\_id" as the key attribute, and all other attributes are appropriately modeled as simple attributes. }
 \correct{The "test\_log" relationship is correctly modeled as a many-to-many (M:N) relationship between "patients" and "test,"}\incorrect{ with "patients" participating totally and "test" participating partially.} 
 The cardinality is correctly set to M for "patients" and N for "test." \correct{The submission does not include any specializations or unions, which aligns with the problem statement. }
 Overall, the submission accurately reflects the entities and relationships described in the problem statement.

\textbf{Mistake 3: remove the total participation constraint+key attribute in test entity+key attribute in patients entity}
\begin{figure}[H]
    \centering
    \includegraphics[width=0.39\textwidth]{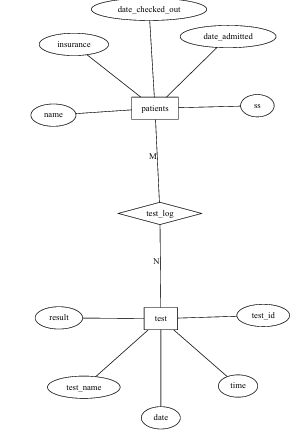}
    \caption{The schema}
    \label{fig:erd_schema}
\end{figure}

\textbf{ASSESSMENT:} GPT-4o clearly exhibits hallucination in this scenario by asserting that "total participation" exists when it does not. Additionally, the model only identifies the entity type without specifying the key attribute that uniquely defines this entity.\\

Gpt-4o: The relationship `test\_log` is correctly identified as a many-to-many (M:N) relationship between `patients` and `test`. The entity `patients` is correctly identified as a strong entity, and the entity `test` is also correctly identified as a strong entity. \incorrect{The participation of `test` in `test\_log` is correctly identified as total,} while the participation of `patients` is correctly identified as partial.
\\

Full R$^2$tA: The feedback evaluates the submitted relationship and entities based on the provided statement and rubrics. \correct{The entities `patients` and `test` are both strong entities, as they have key attributes (`ss` for `patients` and `test\_id` for `test`). However, the submission does not explicitly mark these attributes as keys, which is a potential issue.} The relationship `test\_log` is correctly modeled as a many-to-many (M:N) relationship between `patients` and `test`. Since both entities are strong, the relationship is non-identifying, as it does not propagate keys from one entity to another. \correct{The statement specifies that `test` must totally participate in `test\_log`, meaning every test must be associated with at least one patient, while `patients` can have zero tests, indicating partial participation. The submission does not specify total or partial participation, which is a potential issue.} The feedback should address these points clearly and concisely, ensuring each aspect is evaluated based on the provided rubrics.

\vspace{1em}
\hrule
\vspace{1em}
B+FbSFT+DPO (Baseline w/o reasoning enhancement):
The submission accurately models the entities and relationships as described in the problem statement.
 \incorrect{The "patients" entity is correctly identified as a strong entity with "ss" as its key attribute, and all other attributes are appropriately modeled as simple attributes. }
 \incorrect{Similarly, the "test" entity is correctly identified as a strong entity with "test\_id" as its key attribute,} and all other attributes are appropriately modeled as simple attributes.
\correct{ The "test\_log" relationship is correctly modeled as a many-to-many (M:N) relationship} between \incorrect{"patients" and "test," with "patients" participating totally and "test" participating partially. }
 The cardinality is correctly set to M for "patients" and N for "test." \correct{The submission does not include any specializations or unions, which aligns with the problem statement. }
 Overall, the submission accurately reflects the entities and relationships described in the problem statement.

\textbf{Mistake 4: remove the total participation constraint+key attribute in test entity+key attribute in patients entity+make the relation identifying}
\begin{figure}[H]
    \centering
    \includegraphics[width=0.39\textwidth]{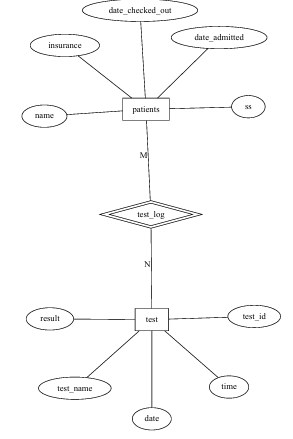}
    \caption{The schema}
    \label{fig:erd_schema}
\end{figure}

\textbf{ASSESSMENT:} GPT-4o is clearly overwhelmed by the volume of errors inserted into the design. In this scenario, it only identifies violations of higher severity while completely missing key attribute-related violations.
\\

Gpt-4o: The relationship `test\_log` is incorrectly identified as an `identifying` relationship.
 According to the problem statement, `test\_log` is a binary many-to-many (M:N) \correct{relationship between `patients` and `test`, which should be non-identifying.}
 \incorrect{The participation of `test` in `test\_log` is correctly identified as total, but the submission does not specify the participation type.}
 The participation of `patients` is partial, which is not explicitly mentioned in the submission.

Full R$^2$tA:
The feedback identifies two main issues with the submitted relationship model: 
1) The relationship type is \correct{incorrectly marked as identifying instead of non-identifying, as both entities are strong }and \correct{should have their own unique keys,however the keys are not specified.}
 2) The \correct{total participation for the `test` entity in the `test\_log` relationship is not specified}, which is required by the problem statement. 
The feedback does not mention any missing relationships or other aspects outside of the submitted model.
\vspace{1em}
\hrule
\vspace{1em}
B+FbSFT+DPO (Baseline w/o reasoning enhancement):

The feedback evaluates the modeling of entities and relationships within a database schema. 
 \incorrect{The entity "patients" is correctly identified as a strong entity, with "ss" serving as its primary key.}
 \incorrect{Similarly, the entity "test" is accurately modeled as a strong entity, with "test\_id" as its primary key.}
 \incorrect{The relationship "test\_log" is appropriately classified as non-identifying}, as neither entity propagates its key into the other. 
 \correct{The cardinality of "M:N" is correctly assigned,} reflecting that each patient can have multiple tests and each test can be associated with multiple patients. 
 However, \incorrect{the participation of "patients" in "test\_log" is incorrectly marked as total;} it should be partial, as not all patients may have tests. 
 Conversely, \incorrect{the participation of "test" in "test\_log" is correctly marked as total}, as every test must be associated with at least one patient. 
\incorrect{ The attributes "ss" for "patients" and "test\_id" for "test" are correctly identified as key attributes}, while "name," "insurance," "date\_admitted," "date\_checked\_out," "test\_name," "date," "time," and "result" are appropriately modeled as simple attributes.

\textbf{Conclusion:} This systematic qualitative analysis reveals compelling insights into LLM behavior when evaluating database schemas, confirming our theoretical predictions and extending prior findings in meaningful ways.

\textbf{Progressive Error Analysis:} Our staged error injection methodology uncovered fascinating behavioral patterns:

\textbf{Mistake 0:} Both models demonstrate competency in recognizing error-free schemas, providing {accurate positive feedback aligned with established rubrics, a promising foundation.

\textbf{Mistake 1 (Critical Violation):} When confronted with a single severe design flaw, both models reliably detect the violation, suggesting robust pattern recognition for fundamental errors.

\textbf{Mistake 2 (+Moderate Violation):} Here the analysis becomes intriguing. With a less critical error in the same relationship, we observe the emergence of our proposed method's reasoning capabilities. The model begins to systematically evaluate each observation before determining correctness, demonstrating structured analytical thinking.

\textbf{Mistake 3 (+Moderate Violation):} GPT-4o begins to exhibit clear hallucinations when overwhelmed by multiple simultaneous mistakes, revealing critical limitations in handling complex error scenarios.

\textbf{Mistake 4 (+Critical Violation):} Most revealing, GPT-4 demonstrates selective discussion, prioritizing critical violations while completely overlooking related key errors,a concerning blind spot for comprehensive evaluation.

\textbf{Broader Implications:} These findings highlight a fundamental challenge: structurally complex data patterns remain underrepresented in standard QA training datasets. Our results suggest that effective evaluation requires chronologically validated conceptual reasoning,as evidenced in the Mistake 2/3/4 scenarios rather than simple pattern matching. The enhanced reasoning framework we propose enriches model cognition through structured cause-and-effect navigation, enabling more sophisticated error detection and extend concerns about LLMs' inability (\ie GPT-4o) to simultaneously capture multiple errors in similar domains~\cite{riazi2025llm}.
\par

To qualitatively examine our findings (complementing the quantitative analysis presented in the main paper), we investigated how our observations align with prior work demonstrating that "fine-tuning alone can degrade chain-of-thought reasoning faithfulness"~\cite{lobo2025impact}.
We analyzed responses from a baseline configuration ("B+FbSFT+DPO") that bypasses reasoning enhancement, applying fine-tuning and DPO alignment directly to feedback without intermediate reasoning steps. When evaluated on multi-mistake scenarios, this baseline exhibited significantly more hallucinations and appeared overwhelmed by complex error patterns, in stark contrast to our reasoning-enhanced approach, which remained faithful to the specification.

\section{Prompt}
\subsection{Relevant Statements Selection Prompt}
We use a GPT-4o prompt to extract relevant descriptions, rubrics, and questions from problem statements for a given  entity-relationship, outputting valid JSON for downstream use in feedback generation.

\vspace{1em}
\lstset{language=prompt}
\begin{lstlisting}
{
    "task": "select the relevant items from the problem statements for explaning the given entity-relationships. make sure the output is a valid json",
    "entity-relationships": {erd},
    "problem-statements": {problem_statements},
    "relevant-statements": [ {"description": "Description from problem-statements.", "rubrics":..., "questions": ...}, ... ]
}
\end{lstlisting}

\subsection{Inference prompt}

The following structured prompt is used to generate feedback on submitted relationships using GPT-4o. Placeholders such as relevant\_statements and submitted\_erd are dynamically filled during execution.

\vspace{1em}

\lstset{language=Prompt}
\begin{lstlisting}
{
  "task": "Provide feedback based on the submitted relationship and participating entities, and their attributes based on the provided solution and problem statements. Each submission only contains one relationship. Do not comment on the missing relationships. Exactly follow the output JSON format.",
  "instructions": [{
    "note": "Only refer to the problem statement to discuss the submission and do not refer to the solution. List all of the entity and attributes discussed in the feedback. Check the submission againts the verifies section in the problem statements",
  }, { "note": "Check the correctness of entity types: Weak and Strong" },{ "note" : "Check the correctness of relationship types: Indentyfying or not identifying"}, { "note" : "Check the total participation has been identified correctly"}, {"note": "Do not comment on the missing relationships"}],
  "context": {
    "statement": "{relevant_statements}",
    "submission": "{submitted_erd}"
  },
  "output": {
    "feedback": "..."
    "entities": [...]
    "attributes": [...]
  }
}
\end{lstlisting}
\vspace{1em}
Please note that for fine-tuning DeepSeek, we leverage reasoning as an intermediate tool to generate better evaluations. This prompt guides the model to produce step-by-step reasoning on the submission, focusing on entity types, relationship types, and participation constraints.
\vspace{1em}
\begin{lstlisting}
{
  "task": "Provide feedback based on the submitted relationship and participating entities, and their attributes based on the provided solution and problem statements. Each submission only contains one relationship. Do not comment on the missing relationships. Exactly follow the output JSON format.",
  "instructions": [{
    "note": "Only refer to the problem statement to discuss the submission and do not refer to the solution. List all of the entity and attributes discussed in the feedback. Check the submission againts the verifies section in the problem statements",
  }, { "note": "Evaluate entity types: Is each entity in the relationship correctly modeled as strong or weak?" },{ "note" : "Check the correctness of relationship types: Indentyfying or not identifying?"}, { "note" : "Check the total participation has been identified correctly"}, {"note": "Do not comment on the missing relationships"}],
  "context": {
    "statement": "{relevant_statements}",
    "submission": "{submitted_erd}"
  },
  "output": {
    "output": {
    "reasoning": "<think>"
  }
}
\end{lstlisting}

\vspace{1em}

This prompt uses the output from the reasoning stage to generate evaluations.
\vspace{1em}
\begin{lstlisting}
{
  "task": "Provide feedback based on the submitted relationship and participating entities, and their attributes based on the provided solution and problem statements. Each submission only contains one relationship. Do not comment on the missing relationships. Exactly follow the output JSON format.",
  "instructions": [{
    "note": "Only refer to the problem statement to discuss the submission and do not refer to the solution. List all of the entity and attributes discussed in the feedback. Check the submission againts the verifies section in the problem statements",
  }, { "note": "Evaluate entity types: Is each entity in the relationship correctly modeled as strong or weak?" },{ "note" : "Check the correctness of relationship types: Indentyfying or not identifying?"}, { "note" : "Check the total participation has been identified correctly"}, {"note": "Do not comment on the missing relationships"}],
  "context": {
    "statement": "{relevant_statements}",
    "submission": "{submitted_erd}",
    "reasoning": "<think>{reasoning}</think>"
  },
  "output": {
    "feedback": {
}
\end{lstlisting}

\subsection{Evaluation prompt}

To evaluate feedback quality from the GPT-4o and our fine-tuned DeepSeek model on  modeling mistakes, we use GPT-4o as an automated judge. For each test instance, a structured prompt analyzes feedback against ground-truth mistakes, classifying detections (true positives), misses (false negatives), hallucinations (false positives), extracting phrases, suggesting ideal feedback, and computing metrics (TP, FN, FP, precision, recall, F1-score). The prompt is as follows:

\vspace{1em}

\lstset{language=JSON}

\begin{lstlisting}
{
  "task": "Evaluate the LLM's detection of known ERD modeling mistakes. For each known mistake type, analyze whether the LLM's feedback or deepseek feedback successfully caught the mistake, even partially. If it missed it, extract any related but incorrect or incomplete explanation. If nothing related exists, leave the field empty. Also suggest what an ideal explanation could have looked like.",
  "instructions": [
    {"note": "The input contains a semicolon-separated list of ground-truth mistake types. These are the actual mistakes in the ERD submission."},
    {"note": "Evaluate both the LLM feedback and deepseek feedback for each mistake type."},
    {"note": "If the mistake is correctly detected, return the relevant sentences from each source that supports detection."},
    {"note": "If the mistake is NOT correctly detected, try to find the relevant sentences from the LLM feedback or deepseek feedback that discusses a related aspect, even if it's incorrect or misapplied."},
    {"note": "If no related discussion exists, return an empty string for that source."},
    {"note": "For each mistake type, add an 'ideal_feedback' field or a short (2 or 3 sentences) example of what a strong, correct conceptual explanation would look like for that mistake in this ERD context."},
    {"note": "Also extract hallucinated mistake claims (false positives) with their relevant sentences, source (LLM feedback or deepseek feedback), and a short explanation."},
    {"note": "Your output MUST be strictly minified JSON without line breaks or indentation. Do not include any escaped characters (\\n, \\t) or unnecessary whitespace within JSON string values. All string values should be plain text without formatting."}
  ],
  "inputs": {
    "focal_relation/entity": "{row['focal_relation']}",
    "mistake_types": "{row['mistake_type']}",
    "num_mistakes": {row['num_mistakes']},
    "problem_statement": "{relevant_statements}",
    "correct_erd": "{correct_erd}",
    "submitted_erd": "{row['mistaken_erd']}",
    "llm_feedback": "{response}",
    "deepseek_feedback": "{deepseek_feedback}"
  },
  "output_format": {
    "mistake_evaluation": [
      {
        "mistake_type": "string",
        "llm_feedback_detected": true,
        "deepseek_feedback_detected": true,
        "llm_feedback_phrase": "phrase from LLM feedback that discusses or detects this mistake (or closest related logic)",
        "deepseek_feedback_phrase": "phrase from deepseek feedback that discusses or detects this mistake (or closest related logic)",
        "ideal_feedback": "a short, ideal feedback sentence that would correctly explain this mistake in this context"
      }
    ],
    "false_positives": [
      {
        "claim_phrase": "string",
        "source": "llm_feedback or deepseek_feedback",
        "why_incorrect": "short explanation"
      }
    ],
    "summary_metrics": {
      "TP_llm_feedback": "int",
      "FN_llm_feedback": "int",
      "FP_llm_feedback": "int",
      "TP_deepseek_feedback": "int",
      "FN_deepseek_feedback": "int",
      "FP_deepseek_feedback": "int",
      "precision_llm_feedback": "float (3 decimals)",
      "recall_llm_feedback": "float (3 decimals)",
      "f1_score_llm_feedback": "float (3 decimals)",
      "precision_deepseek_feedback": "float (3 decimals)",
      "recall_deepseek_feedback": "float (3 decimals)",
      "f1_score_deepseek_feedback": "float (3 decimals)"
    }
  }
}
\end{lstlisting}

\subsection{Data generation prompt}

We utilize a prompting-based approach with GPT-4o to systematically introduce realistic, student-like errors into a correct entity-relation diagram while preserving syntactic validity. This method ensures diversity and balance across predefined mistake categories (e.g., keys, participation constraints, entity types), generating semantically incorrect but grammatically valid. The process operates in batches, dynamically prioritizing underrepresented mistake types to achieve balanced coverage. Each generated instance focuses on one focal relation or entity, introducing a varying number of mistakes from applicable categories based on the structure, without adding new elements.
The generation is driven by a structured prompt that includes: (\romannum{1}) task requirements emphasizing syntactic validity and mistake constraints; (\romannum{2})  a detailed  grammar specification for entities, relationships, specializations, and unions; (\romannum{3})  an example correct diagram; (\romannum{4})  targeted mistake categories for the batch; and (\romannum{5})  output format as JSON for easy parsing. The prompt is instantiated per batch with the correct diagram grammar, batch size, starting ID, and least-used mistake types (selected via a counter to promote balance).\par

To ensure our method can effectively handle  varying numbers of mistakes, we generate datasets that include instances with single, double, and triple mistakes. Specifically, for each  schema, we create 50 instances with one mistake, 50 with two mistakes, and 50 with three mistakes, resulting in a total of 150 instances per schema.\par

The following prompt template is used to generate mistaken variants using GPT-4o. Placeholders such as batch\_size, start\_index,number\_of\_errors and mistake categories are dynamically filled during execution to ensure balanced data generation. 

\vspace{3em}

\lstset{language=Prompt}
\begin{lstlisting}
You will create multiple incorrect versions of a correct Entity-Relationship Diagram (ERD) by introducing realistic student mistakes while maintaining syntactic validity (based on ERD Grammar).

Task Requirements:
- Focus on ONE relation or entity per mistaken ERD from the provided correct ERD.
- Introduce 2 realistic, student-like errors in the chosen relation/entity.
- IMPORTANT: DO NOT create new entities or relationships that does not exist in the correct ERD.
- ONLY modify the provided ERD by introducing errors to EXISTING entities and relationships.
- First analyze the ERD structure to understand what elements are present (entities, relationships, specializations, etc.).
- Only introduce error types that are applicable to the structures that actually exist in the diagram.
- For example, only apply ternary relationship errors if the ERD contains ternary relationships.
- Maintain valid ERD grammar (no syntax violations).
- Ensure mistakes are semantically incorrect but syntactically valid.
- The complete ERD should be identical to the correct ERD except for the specific errors introduced.
- Output in the specified JSON format.

[ERD Grammar]

Correct ERD:
[Correct ERD]

For this batch, focus on creating mistakes primarily from these categories:

- {category1}: {description1}. Specific error types to consider: {subtypes1}
- {category2}: {description2}. Specific error types to consider: {subtypes2}
- {category3}: {description3}. Specific error types to consider: {subtypes3}

IMPORTANT: First analyze the ERD to determine which error types are applicable. Only use error types that make sense for the given ERD structure. For example, only apply ternary relationship errors if the ERD actually contains ternary relationships.

You MUST create {batch_size} different mistaken ERDs by introducing EXACTLY {number_of_errors} errors into the EXISTING entities and relationships in the correct ERD. Each mistaken ERD should have EXACTLY {number_of_errors} mistakes, no more and no fewer. DO NOT create new entities or relationships. Start the mistake_id numbering at {start_index}.

Output Format:

{
  "mistaken_erds": [
    {
      "mistake_id": 1,
      "focal_relation": "RelationName",
      "description": "Brief description of the mistake(s)",
      "mistakes": [
        {
          "type": "error_type_1",
          "original": "Original correct structure",
          "modified": "Modified incorrect structure"
        },
        {
          "type": "error_type_2",
          "original": "Original correct structure",
          "modified": "Modified incorrect structure"
        }
      ],
      "mistaken_erd": "The complete ERD in textual syntax format, identical to the 'Correct ERD' except with the specified mistake(s) applied. The structure must include ALL entities, relationships, specializations, and unions from the original correct ERD, with only the specific modifications for the mistakes."
    }
  ]
}
\end{lstlisting}

\subsection{ Enhancement of reasoning}

We employ a self-reflective refinement loop to polish both initial LLM-extracted reasoning and associated feedback on modeling mistakes by addressing false positives (hallucinated claims) and false negatives (missed mistakes). The process evaluates outputs against ground-truth using GPT-4o, removes erroneous phrases via string replacement, inserts ideal explanations for omissions, and iterates (up to  5 attempts) until precision and recall reach 1.0 or attempts are exhausted. Refinement occurs in parallel branches: one for feedback (\texttt{polished\_feedback}) and one for reasoning (\texttt{polished\_feedback\_reasoning}), with metrics tracked separately (e.g., \texttt{refined\_precision} vs. \texttt{refined\_precision\_wreason}). This enhances accuracy and reliability.

The pipeline leverages the Feedback Generation Prompt (see Section 4.2 for details) for initial outputs and the Evaluation Prompt (detailed in Section 4.3) for assessment, ideal explanations, and metric computation.

\end{document}